\documentclass[a4paper, 10pt, conference]{ieeeconf}      

\IEEEoverridecommandlockouts                              

\usepackage{float}

\usepackage{graphicx} 

\usepackage{todonotes}
\usepackage{silence}
\usepackage{pgfplots} 
\pgfplotsset{compat=1.16}

\usepackage{url}

\usepackage[acronym]{glossaries}
\newacronym{IR}{IR}{intention recognition}

\title{\LARGE \bf Uncertainty-Resilient Active Intention Recognition\\ for Robotic Assistants}

\author{
Juan Carlos Saborío$^{2,1,*}$,
Marc Vinci$^{1,3,*}$,
Oscar Lima$^{1,3}$,
Sebastian Stock$^{1}$,
Lennart Niecksch$^{1,3}$,\\
Martin Günther$^{1}$,
Alexander Sung$^{1}$,
Joachim Hertzberg$^{1,3}$ and
Martin Atzmueller$^{1,3}$
\thanks{This work is supported by the InCoRAP, ExPrIS and LIEREx projects through grants from the German Federal Ministry of Education and Research (BMBF) with Grant Numbers 0IW19002, 01IW23001 and 01IW24004. The DFKI Niedersachsen (DFKI NI) and the Joint Lab for AI \& DS are sponsored by the Ministry of Science and Culture of Lower Saxony and the VolkswagenStiftung.}%
\thanks{$^{1}$ German Research Center for Artificial Intelligence (DFKI), Osnabrück, Germany {\tt\small \{marc.vinci, oscar.lima, sebastian.stock, lennart.niecksch, martin.guenther, alexander.sung, joachim.hertzberg, martin.atzmueller\}@dfki.de}}%
\thanks{$^{2}$ Osnabrück University, Joint Lab for AI \& DS, Osnabrück, Germany {\tt\small jcsaborio@uos.de}}%
\thanks{$^{3}$ Osnabrück University, Institute of Computer Science, Osnabrück, Germany}%
\thanks{$^{*}$ These authors contributed equally to this publication.}%
}

\begin{document}
\maketitle
\thispagestyle{empty}
\pagestyle{empty}

\begin{abstract}
Purposeful behavior in robotic assistants requires the integration of multiple components and technological advances. Often, the problem is reduced to recognizing explicit prompts, which limits autonomy, or is oversimplified through assumptions such as near-perfect information. We argue that a critical gap remains unaddressed -- specifically, the challenge of reasoning about the uncertain outcomes and perception errors inherent to human intention recognition.  In response, we present a framework designed to be resilient to uncertainty and sensor noise, integrating real-time sensor data with a combination of planners. Centered around an intention-recognition POMDP, our approach addresses cooperative planning and acting under uncertainty. Our integrated framework has been successfully tested on a physical robot with promising results.
\end{abstract}

\section{Introduction}
Robotic assistants may be integrated into modern industrial environments, e.g., delivering tools, parts or modules interleaved with tidying the workspace. Such tasks, however, require a combination of robust planning, navigation, grasping, and perception--particularly when explicit commands are not available and the robot must identify and pursue goals, in collaborative spaces shared with people.

Existing work already addresses several building blocks necessary for such behavior, ranging from flexible robot control \cite{Mavsar2021,Gottardi2023} to \gls{IR} and partially observable planning \cite{RamirezGeffner2011, Amato2019, Cramer2021, SaborioHertzbergICAPS2023}. \gls{IR} based on Partially Observable Markov Decision Processes (POMDPs) provides a formal framework for handling uncertainty and information gathering, but a gap remains when transfering these methods to real robotic platforms. Recent approaches include detecting motion direction \cite{Mavsar2021} or recognizing explicit gestures that trigger replanning \cite{Gottardi2023}, but both are limited to \emph{activities} rather than predicting future needs, failing to address the inherent uncertainty in \gls{IR} which stems from both perception errors and the unpredictability of actions with delayed outcomes.

Motivated by a human-robot collaboration scenario, we propose an integrated framework that tackles several open challenges in automated robotic assistance, with a focus on uncertainty and \gls{IR}. Key challenges include dynamic environments, incomplete information, and the need to leverage real-time sensor data for online action selection.  The collaborative aspect is addressed through human \emph{\gls{IR}}: the ability of an agent to infer and predict another's upcoming actions or needs. This interpretation assumes agents behave in a goal-directed manner, where an intention refers to a potential (sub)goal or future action within their plan \cite{FreedmanZilberstein2017}, rather than a currently ongoing activity. Acting based on predicted intentions enables autonomy that goes beyond avoiding collisions or reacting to explicit instructions, such as spoken commands or gestures.

We implemented and tested our approach through integrated system comprising external cameras, a mobile robot, and multiple software modules for data acquisition, processing, task selection, and action execution. At its core is an \emph{active} goal recognition (AGR) POMDP model and planner, which integrates external information and selects the next best action, corresponding to a high-level task further decomposed and executed onboard the robot by a hierarchy of planners.

\begin{figure}
    \centering
    \includegraphics[width=0.5\linewidth]{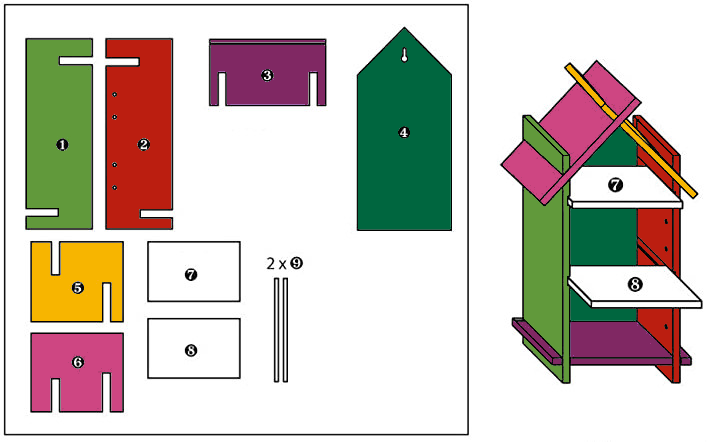}
    \caption{A ten-part insect hotel: individual parts (left) and assembled (right).}
    \label{fig:insect hotel}
\end{figure}

Each module was validated independently, and the integrated framework was tested on a physical robot in a simplified industrial assembly scenario. In our tests, a human worker assembles one of two types of \emph{insect hotels}, small wooden structures designed to shelter insects (see Fig.~\ref{fig:insect hotel}). Each hotel type requires both common and type-specific parts, all color-coded, that may be assembled in different orders, generalizing the six-part ball point pen used in \cite{Cramer2021}. The robot supports the worker by monitoring part usage and availability and by delivering missing parts \emph{as needed}. Our system was able to assist a human worker effectively despite uncertainty and sensor noise and without relying on explicit instructions.

The main contribution of this paper is the integrated framework for uncertainty-resilient active \gls{IR} in robotic assistants, focusing on information and outcome uncertainty in cooperative settings. We start by reviewing related work in \gls{IR} with robotic applications, followed by a detailed description of our framework, evaluated on two sets of simulated experiments. We conclude with a brief discussion of current limitations and directions for future work.


\section{Related Work}
\textbf{Intention Recognition in HRC.}
Recent advances in \gls{IR} for human-robot collaboration (HRC) adopt a reactive rather than \emph{proactive} perspective. Examples include motion and gesture recognition to trigger HTN replanning \cite{Gottardi2023}, and an RNN approach to detect motion and trajectories to avoid interrupting or disrupting an ongoing task \cite{Mavsar2021}.  We address this gap in proactive assistance by recognizing needs or goals, and not merely ongoing actions.
Another approach models human actions as an ``intention graph'' and uses an offline POMDP planner for \gls{IR} and action selection \cite{Cramer2021}, but assumes the robot performs the same actions as a human, on the same object by taking turns with no synchronization or coordination steps. Furthermore no robots were actually involved and there is no mention of physical limitations, failures or delays. We use our own extended POMDP formulation \cite{SaborioHertzbergICAPS2023}, that allows a physical robot to reason about its own actions and their effect on a (human) worker, coordinating their behavior in a shared cooperative space.  The resulting large POMDP is solved online, without prior factorization, by integrating noisy sensor information into an optimization process supporting joint planning and execution.

\textbf{From Perception to Intention.}
Recent approaches in HRC have achieved strong performance in \emph{activity} recognition, based on gestures and ongoing tasks using skeleton-based tracking \cite{Terreran2023}, hand-object contact modeling \cite{Pan2023}, or HMMs \cite{Wang2020}.
We instead focus on generating high-level abstractions (e.g., state and subgoal conditions) from multimodal sensor data, that become informative observations for the POMDP planner.
In our demo scenario, we combined RGB-D cameras with a trained YOLOv8~\cite{YoloV8} to detect relevant parts and track task progress, but any setup is feasible as long as similar abstractions can be generated.

\textbf{Intention, Plan and Goal Recognition.}
IR is traditionally restricted to observing and classifying the behavior of an agent \cite{RamirezGeffner2010, RamirezGeffner2011, Geib2015, Sukthankar2014}, relying on plan libraries with precomputed plans or a grammar to construct them \cite{GeibGoldman2009, Geib2015, Rafferty2017} or through reverse planning methods that expand the space of plans and approximate possible goals \cite{RamirezGeffner2010,FreedmanZilberstein2017}, also covering the cases of missing information \cite{Sohrabi2016} and partially observable domains \cite{RamirezGeffner2011}.
We build upon these advances with a focus on planning and acting \emph{online}, important in large and complex domains \cite{MassardiBeaudry2021}, and also by adopting the \emph{active} approach \cite{Amato2019}, where the robot is not just a passive observer but a participant with its own actions. Furthermore, we don't rely on state factoring or other types of preprocessing to simplify the domain \cite{SaborioHertzbergICAPS2023}.

\textbf{POMDP Planning in \gls{IR}.}
POMDP-based \gls{IR} approaches in robotics often rely on offline, point-based solvers \cite{Cramer2021, MassardiBeaudry2021}, limited to small problems and factored state representations.
We instead use RAGE, a POMDP planner that supports online planning via relevance estimation and subgoal generation \cite{SaborioHertzbergLNAI19, SaborioHertzbergUAI19}.
These extensions to Monte-Carlo planning improve performance in \gls{IR} problems and complex POMDPs, necessary to successfully interleave perception, planning and acting. Reference planners such as POMCP \cite{SilverPOMCP} can solve our proposed model but struggle with \gls{IR} challenges such as delayed rewards. Such recent advances provide a fresh perspective on how to integrate POMDP methods onboard robots.

\textbf{Robot Control Architectures for HRC.}
Several HRC systems rely on specialized controllers that integrate multi-modal perception and reactive planning \cite{GeaFernandez2017ras}, or explore structured control for human-robot interaction (HRI) and collaborative manipulation \cite{Paxton2017, Alami2006}.
Recent systems that integrate task and motion planning (TAMP) also interleave symbolic task-level reasoning with continuous motion planning \cite{Dantam2016tamp}, enabling robots to adapt to environmental constraints and task changes.
These systems focus on deterministic environments and do not explicitly address behavioral uncertainty or perceptual ambiguity, whereas our system tightly couples planning and control under uncertainty via online POMDP planning, allowing the robot to act proactively and recover from failures. Highly experimental alternatives include grounded vision-language-action models \cite{zawalskiVLA}, but planning performance is severely limited even in simple scenarios with only immediate action outcomes.

\textbf{Application Domains.}
Collaborative task execution with intention-aware robots is especially relevant in manufacturing domains, where plans are typically semi-structured and multiple agents interact under safety and performance constraints \cite{Schlenoff2015}.
The insect hotel problem captures these dynamics and also highlights broader HRI challenges in communication, turn-taking, and shared decision-making \cite{Zhang2023}.

\section{Active Goal Recognition with Robots}
\label{sec:Overview}
\begin{figure*}
    \centering
    \includegraphics[width=0.7\linewidth]{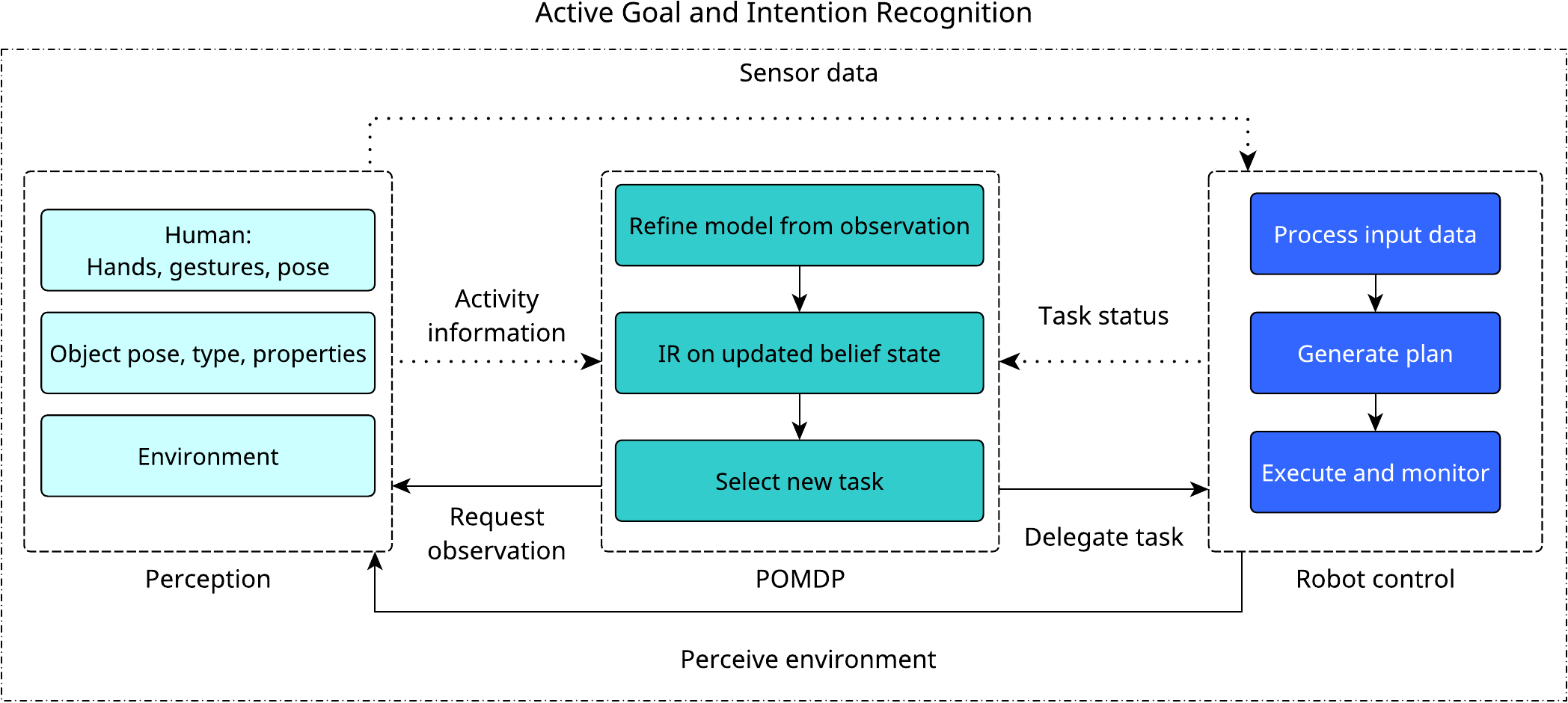}
    \caption{AGR Architecture: Dashed lines group the different major modules and arrows indicate the necessary communication and representation steps between them. Solid arrows represent actions, while dotted arrows represent action outcomes, data, etc. The POMDP planner selects high-level tasks implemented on the hardware following \ref{sec:Perception} for information gathering, and \ref{sec:ExecutionMonitoring} for manipulation. The robot itself also uses perception functions, some of which are executed using onboard sensors.}
    \label{fig:AGR_PBRC}
\end{figure*}

We propose an architecture, illustrated in fig.~\ref{fig:AGR_PBRC}, that combines perception, representation, high- and low-level planning, and action execution onboard a physical robot, highlighting the essential role of modular integration in the system's performance.

The perception subsystem includes recognition of human-related features such as hand positions, gestures, and body pose, activities and their outcomes as well as properties of inanimate objects and, potentially, changes in the environment.  
Perception actions are coordinated by the POMDP planner, and their outcome must be provided in a format expected by the planner such as symbolic labels or feature vectors, so they can be incorporated into the model as observations. 

The POMDP subsystem performs high-level (active) intention recognition and goal reasoning, selecting actions in response to perceived events, human activities and object properties.
During operation, the POMDP planner approximates a control policy informed by activity observations and selects relevant tasks that can fulfill goals. Information-gathering tasks are performed by the perception subsystem, while navigation and manipulation tasks are delegated onto the robotic platform.

The functions of the robot control software stack are encapsulated in the right-most part of the diagram, which includes high-level task planning and plan execution, and low-level navigation, localization, movement and velocity control, arm and gripper motion planning, some of which may require additional perception steps.
Plans are executed by calling the corresponding action implementations on the robot, and information about plan and action outcomes is communicated back to the POMDP planner to update the internal belief state. The planner can now proceed searching for a new task with updated information.

In the diagram, the arrows illustrate the various communication and representation steps required to transform and send data, such as converting the output of an object and gesture classifier into a labeled activity, or dispatching and executing a high-level task (selected by the POMDP planner) on a mobile robot.

\section{System Description}
Below we describe the various modules and implementation challenges, using our chosen application domain as a guiding example.

\subsection{Insect Hotel Assembly}
\label{sec:InsectHotel}
An insect hotel is assembled by selecting parts from an inventory area and placing them in designated sections of a workspace. Necessary parts may be missing so the robot must closely monitor events and respond \emph{as needed}. The challenge lies in correctly understanding the plan pursued by the worker, since hotels may follow different sequences of steps and require some unique parts. For simplicity we use color-coded parts and rely on color recognition. The robotic assistant continually estimates part availability, part assembly status and hotel type through the AGR-POMDP, and generates suitable tasks in response.

\begin{figure}
    \centering
    \includegraphics[width=\linewidth]{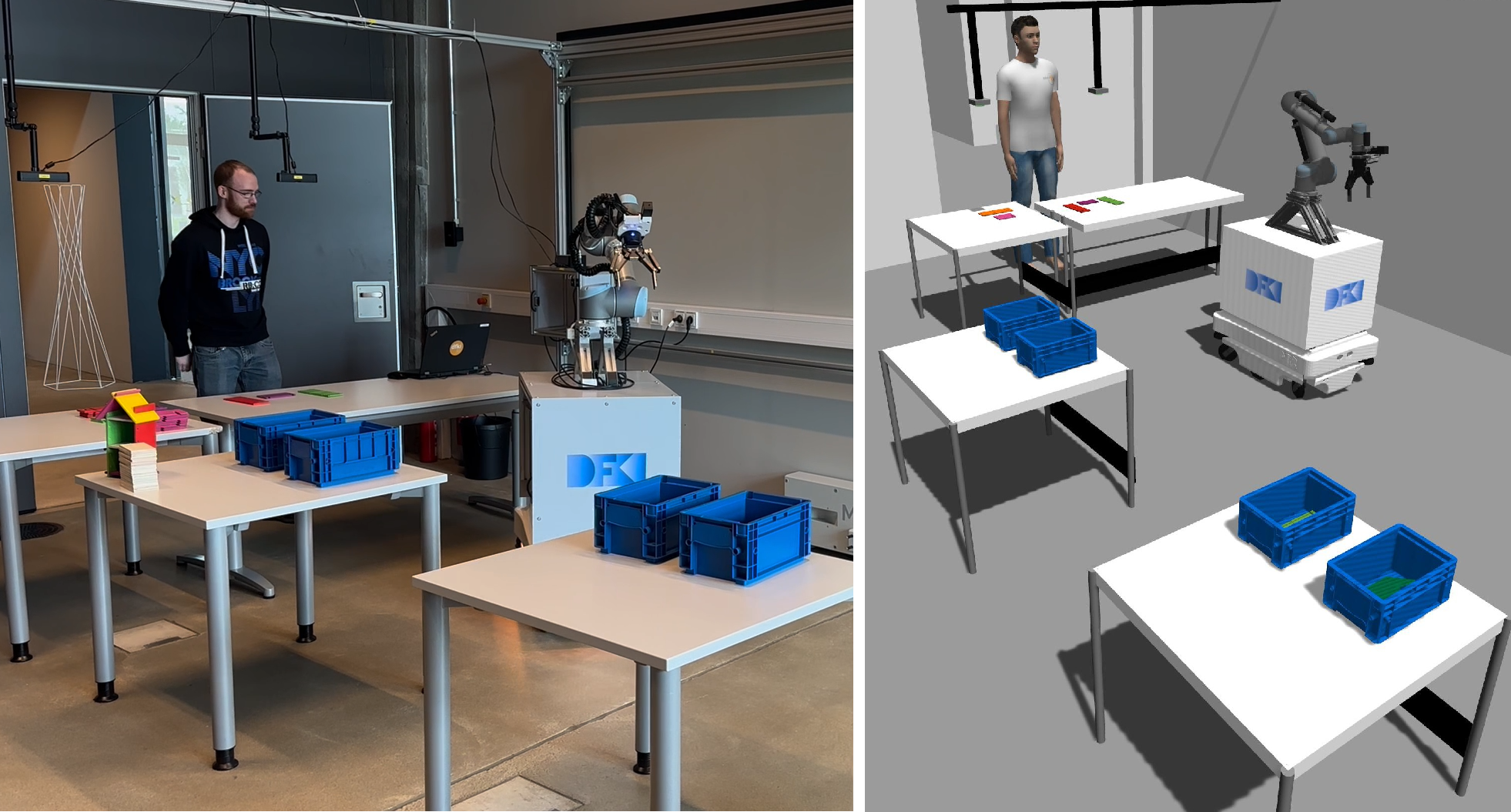}
    \caption{Robot-assisted assembly environment, with the physical Mobipick robot (left) and in Gazebo simulation (right). Back: inventory area (smaller table) and workspace (larger table), both holding insect hotel parts; Front: storage area with parts in boxes.}
    \label{fig:demo_scenario_photo}
\end{figure}

We set two overhead cameras, one pointed at the inventory and one at the workspace, used to generate observations. The worker may pick any available part and attempt to assemble it, and may even make and fix mistakes. The different possible orders constitute the space of plans and the hotel types, the goals.
The scenario with a physical robot and a human worker was replicated in the Gazebo simulator for controlled experimentation (Fig.~\ref{fig:demo_scenario_photo}). Parts are placed in a storage area and stored in boxes that can be easily grasped and carried over to the worker.

\subsection{Perception and object detection}
\label{sec:Perception}


We implemented a simplified version of the perception subsystem by combining human activity and object perception, reducing this to detecting color-coded parts. Detailed human activity recognition would improve the overall performance and versatility, but it was out of the scope of our projects.

Our system recognizes parts, boxes and box contents, which helps monitor part availability, assembly progress, infer completed worker actions and reason about worker intentions.
We also detect boxes in the storage area, identify their contents and estimate their 6DoF pose with sufficient precision for grasping and delivery.

Part detection is supported by two ASUS Xtion cameras that feed a YOLOv8~\cite{YoloV8} model, trained on a synthetic dataset generated from 3D meshes.
This dataset was created using NVISII~\cite{morrical2021nvisii} and contains the relevant and distractor objects, rendered in front of different High Dynamic Range (HDR) domes as backgrounds.
\footnote{The synthetic dataset and the trained YOLOv8 weights are available at \url{https://doi.org/10.5281/zenodo.15190122}}
For 6DoF pose estimation of the boxes, we deployed a DOPE~\cite{tremblay2018corl:dope} model on the Mobipick robot.
Additionally, a simple HSV-based color classifier was used for the box contents.
Currently, we are working on a unified approach for object detection, 6D pose estimation and inference of spatial relations that can replace these separate systems \cite{niecksch2024meshbasedobjecttrackingdynamic}.

The same perception pipeline may be used in Gazebo, but for better control under simulation we evaluated using ground-truth data (Sec.~\ref{sec:results}), with artificial sensor noise.



\subsection{Intention recognition and goal selection}
\label{sec:IR}
Plan approximation and action prediction was modeled as an AGR-POMDP \cite{SaborioHertzbergICAPS2023}.\footnote{Robot and human states, actions, etc. follow an equivalent formulation.} The worker's task model is an MDP policy that can generate plans and actions, such as assembling or removing parts, taking pauses and even making mistakes, all with probability. Through the AGR-POMDP the robot estimates if parts are available or assembled (via observations), and selects tasks (the solid arrows in Fig.~\ref{fig:AGR_PBRC}) that help with hotel assembly.

Perception tasks are implemented as per Sec.~\ref{sec:Perception}, leading to labeled observations. Manipulation tasks are delegated to the robot platform and return a status signal, as explained in Sec.~\ref{sec:ExecutionMonitoring}. The resulting system deliberates about worker and robot actions, supported by real-world sensor data, and provides assistance without the need for explicit prompts such as commands or gestures.

If objects are detected incorrectly or grasping is unsuccessful, the POMDP planner responds by gathering more observations or reevaluating the cost-benefit ratio of grasping again, without hard-coded behaviors or arbitrary conditions. We argue this is a much more natural and flexible approach to handle inherent uncertainty.

Delayed rewards are a significant challenge in AGR, also handled by our system. The robot is required to select actions in anticipation of predicted events so their true benefit can only be assessed afterwards.
The underlying AGR-POMDP may be solved with any well-suited planner (such as POMCP, c.f. sec. \ref{sec:results}), but our results show a substantial benefit when implementing relevance estimation \cite{SaborioHertzbergLNAI19,SaborioHertzbergUAI19}.

\subsection{Robot control, action execution and monitoring}
\label{sec:ExecutionMonitoring}
Tasks selected by the POMDP planner are given to the robot control system as new goals, pursued through the integration of several modules including HTN-planning from the Unified Planning library~\cite{micheli2025unified}, dependency-graph-based plan execution based on the Embedded Systems Bridge~\cite{sadanandam2023closed} as well as the low-level robot capabilities provided by ROS, i.e., move\_base, MoveIt and a custom grasp planner.\footnote{\url{https://github.com/DFKI-NI/grasplan}} For further infrastructure and hardware details see \cite{Lima2023icaps}.
The hierarchical task planner decomposes incoming goals into subtasks, generating a feasible plan with only primitive actions.
These actions are sequentially dispatched on the robot with a plan executor that calls corresponding ROS action implementations, and reports the result of these actions (and plan) to the POMDP subsystem to continue planning on an updated state.

The tight interplay between perception, goal recognition, planning and action execution leads to rich, emerging dynamics that balance the cost and opportunity of acting under uncertainty, with noisy information and perception errors.




\section{Results}
\label{sec:results}
Our system runs onboard a physical robot,\footnote{See \url{https://dfki-ni.github.io/IR_ECMR_2025} for a detailed video and demo source code.} but for replicability we rely on two sets of simulated experiments. In the first, we tested the resiliency of the AGR-POMDP against uncertainty, perception errors and sensor noise, and in the second we tested the dynamics of robotic assistance in a Gazebo simulation.


\subsection{\gls{IR} and sensor noise}
In the AGR-POMDP perception actions yield a reward of $-0.5$, while restocking parts award $-10$ without enough information, $2$ under ideal conditions (information, availability, etc.) and $-2$ otherwise. The worker task model generates rewards for missing parts ($-2$), successful assembly ($2$) and completed objects ($5$). This seemingly simple problem is in fact quite challenging for an AGR robot, which must track a moving target using incomplete information, and also predict what it might do next.

We measured performance on a set of smaller hotels consisting of 3 common parts plus 1 type-specific part each, with random initial inventory and hotel types and increasing levels of sensor accuracy. The results average discounted returns and standard errors over 100 runs under different planning budgets, using the relevance-based RAGE planner as well as POMCP (uniformly random sampling) as a standard baseline, with a maximum of 100 steps and a discount factor of $\gamma = 0.99$.

\begin{figure}[ht]
    \centering

\definecolor{acc50}{rgb}{0.6,0.0,0.0}
\definecolor{acc65}{rgb}{0.8,0.4,0.0}
\definecolor{acc75}{rgb}{0.1,0.5,0.1}
\definecolor{acc85}{rgb}{0.0,0.0,0.8}

\pgfplotscreateplotcyclelist{mbw}{%
solid, every mark/.append style={solid, fill=white}, mark=triangle\\%
solid, every mark/.append style={solid, fill=white}, mark=square*\\%
solid, every mark/.append style={solid, fill=white}, mark=*\\%
solid, every mark/.append style={solid, fill=white}, mark=star\\%
solid, every mark/.append style={solid, fill=acc50}, mark=triangle\\%
solid, every mark/.append style={solid, fill=acc65}, mark=square*\\%
solid, every mark/.append style={solid, fill=acc75}, mark=*\\%
solid, every mark/.append style={solid, fill=acc85}, mark=star\\%
}

\begin{tikzpicture}[scale=0.8]
\begin{semilogxaxis}[
	xlabel={Simulations},
	ylabel={Avg. Discounted Return},
	cycle list name=mbw,
	legend style={at={(0.5,-0.2)},anchor=north},
	legend columns=4
]

\addplot+[acc50, dashed, error bars/.cd,y dir=both,y explicit, error bar style={solid}]
coordinates {
(2,-95.99) +- (0,3.75)
(4,-92.38) +- (0,3.476)
(8,-77.15) +- (0,3.085)
(16,-74.96) +- (0,2.723)
(32,-75.64) +- (0,3.01)
(64,-66.8) +- (0,2.919)
(128,-61.06) +- (0,2.067)
(256,-54.98) +- (0,1.95)
(512,-49.41) +- (0,1.592)
(1024,-44.31) +- (0,1.5)
(2048,-42.22) +- (0,1.259)
(4096,-42.67) +- (0,1.097)
(8192,-40.75) +- (0,1.191)
(16384,-40.63) +- (0,1.189)
(32768,-40.93) +- (0,1.102)
(65536,-42.06) +- (0,1.096)
};

\addplot+[acc65, dashed, error bars/.cd,y dir=both,y explicit,  error bar style={solid}]
coordinates{
(2,-96.34) +- (0,3.802)
(4,-93.83) +- (0,4.178)
(8,-75.83) +- (0,3.408)
(16,-69.11) +- (0,3.251)
(32,-64.29) +- (0,2.938)
(64,-69.6) +- (0,3.024)
(128,-59.26) +- (0,2.965)
(256,-52.96) +- (0,1.722)
(512,-47.1) +- (0,1.503)
(1024,-40.78) +- (0,1.153)
(2048,-41.74) +- (0,1.337)
(4096,-38.8) +- (0,1.184)
(8192,-39.59) +- (0,1.08)
(16384,-37.93) +- (0,1.134)
(32768,-38.04) +- (0,1.143)
(65536,-38.31) +- (0,1.122)
};

\addplot+[acc75, dashed, error bars/.cd,y dir=both,y explicit,  error bar style={solid}]
coordinates{
(2,-101.9) +- (0,4.475)
(4,-95.04) +- (0,4.059)
(8,-74.33) +- (0,3.267)
(16,-76.63) +- (0,3.46)
(32,-64.95) +- (0,2.582)
(64,-62.84) +- (0,2.828)
(128,-52.17) +- (0,1.987)
(256,-44.79) +- (0,1.685)
(512,-39.45) +- (0,1.617)
(1024,-35.48) +- (0,1.552)
(2048,-35.8) +- (0,1.373)
(4096,-34.96) +- (0,1.515)
(8192,-33.11) +- (0,1.275)
(16384,-35.49) +- (0,1.391)
(32768,-32.54) +- (0,1.413)
(65536,-34.03) +- (0,1.34)
};

\addplot+[acc85, dashed, error bars/.cd,y dir=both,y explicit,  error bar style={solid}]
coordinates{
(2,-91) +- (0,3.276)
(4,-89.36) +- (0,3.7)
(8,-76.52) +- (0,3.081)
(16,-78.93) +- (0,3.204)
(32,-68.99) +- (0,2.958)
(64,-66.39) +- (0,2.733)
(128,-52.66) +- (0,2.083)
(256,-44.21) +- (0,1.708)
(512,-39.1) +- (0,1.721)
(1024,-37.16) +- (0,1.301)
(2048,-32.39) +- (0,1.276)
(4096,-32.97) +- (0,1.371)
(8192,-30.63) +- (0,1.326)
(16384,-31.77) +- (0,1.387)
(32768,-32.53) +- (0,1.476)
(65536,-30.26) +- (0,1.442)
};

\addplot+[acc50, solid, error bars/.cd,y dir=both,y explicit,  error bar style={solid}]
coordinates{
(2,-96.53) +- (0,4.019)
(4,-87.62) +- (0,3.618)
(8,-75.77) +- (0,2.883)
(16,-74.81) +- (0,3.244)
(32,-64.42) +- (0,2.854)
(64,-62.31) +- (0,3.144)
(128,-49.12) +- (0,2.618)
(256,-41.46) +- (0,2.07)
(512,-33.23) +- (0,2.012)
(1024,-27.38) +- (0,1.674)
(2048,-29.33) +- (0,1.623)
(4096,-28.19) +- (0,1.859)
(8192,-26.91) +- (0,1.531)
(16384,-28.67) +- (0,1.605)
(32768,-28.13) +- (0,1.542)
(65536,-28.44) +- (0,1.645)
};

\addplot+[acc65, solid, error bars/.cd,y dir=both,y explicit,  error bar style={solid}]
coordinates{
(2,-99.29) +- (0,4.6)
(4,-84.18) +- (0,3.762)
(8,-78.18) +- (0,2.901)
(16,-63.46) +- (0,2.822)
(32,-61.73) +- (0,2.839)
(64,-57.39) +- (0,2.757)
(128,-44.53) +- (0,2.324)
(256,-35.28) +- (0,1.557)
(512,-26.84) +- (0,1.468)
(1024,-23.2) +- (0,1.574)
(2048,-26.98) +- (0,1.484)
(4096,-25.68) +- (0,1.41)
(8192,-29.2) +- (0,1.454)
(16384,-27.03) +- (0,1.282)
(32768,-25.4) +- (0,1.515)
(65536,-26.6) +- (0,1.475)
};

\addplot+[acc75, solid, error bars/.cd,y dir=both,y explicit,  error bar style={solid}]
coordinates{
(2,-96.04) +- (0,4.117)
(4,-78.39) +- (0,3.175)
(8,-73.27) +- (0,2.765)
(16,-66.2) +- (0,3.204)
(32,-58.08) +- (0,2.836)
(64,-44.8) +- (0,2.519)
(128,-41.09) +- (0,2.329)
(256,-30.14) +- (0,1.584)
(512,-21.7) +- (0,1.563)
(1024,-20.04) +- (0,1.496)
(2048,-12.2) +- (0,1.308)
(4096,-13.22) +- (0,1.408)
(8192,-12.68) +- (0,1.186)
(16384,-11.99) +- (0,1.079)
(32768,-10.25) +- (0,1.161)
(65536,-10.99) +- (0,1.252)
};

\addplot+[acc85, solid, error bars/.cd,y dir=both,y explicit,  error bar style={solid}]
coordinates{
(2,-96.92) +- (0,3.969)
(4,-85.44) +- (0,3.512)
(8,-74.27) +- (0,2.952)
(16,-70.53) +- (0,2.975)
(32,-57.88) +- (0,2.532)
(64,-48.2) +- (0,2.148)
(128,-35.53) +- (0,1.916)
(256,-27.05) +- (0,1.545)
(512,-23.09) +- (0,1.221)
(1024,-16.43) +- (0,1.269)
(2048,-9.004) +- (0,1.105)
(4096,-9.074) +- (0,1.074)
(8192,-8.993) +- (0,1.16)
(16384, -10.7) +- (0,1.18)
(32768, -11.96) +- (0,1.232)
(65536,-9.893) +- (0,1.125)
};

\legend{POMCP 0.5, POMCP 0.65, POMCP 0.75, POMCP 0.85, RAGE 0.5, RAGE 0.65, RAGE 0.75, RAGE 0.85}
\end{semilogxaxis}
\end{tikzpicture}
    \caption{Active Goal Recognition under different levels of sensor accuracy, using the POMCP and RAGE planners. A sensor accuracy of $0.5$ is a lower-bound representing maximum uncertainty. The range of mean returns is problem-dependent.}
    \label{fig:PlanningPerformance}
\end{figure}
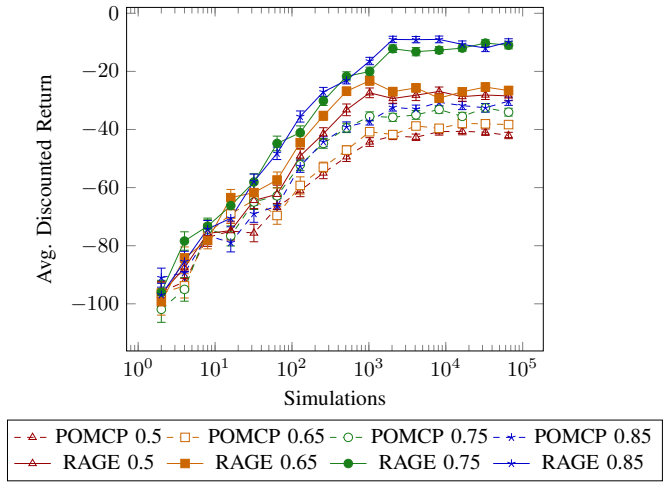

Figure~\ref{fig:PlanningPerformance} shows that both POMDP planners are capable of maintaining a baseline performance despite extreme sensor uncertainty, and increase their returns with a more generous planning budget. Starting at only 256 simulations per step, both planners consistently reach terminal states (i.e., fully assembled hotels) even with very low perception accuracy, and increase their performance with only small improvements in information quality. Better sensor accuracy results in more reliable observations, leading to fewer mistakes and more accurate predictions. As expected, we also perceived substantially better returns with RAGE, highlighting the need for sophisticated approaches in AGR.

\subsection{Assistance scenario}
We replicated several exemplary scenarios in Gazebo, using 2 hotel types each with a total of 6 parts: 4 common, and 2 type-specific: red and dark green for type-A, orange and black for type-B.
In the various scenarios, the robot generally avoids unique parts until there is enough information and their cost is well justified. This means that, albeit with potential delays, the robot pursues tasks that respond to the worker's behavior.

\begin{figure*}
    \centering
    \includegraphics[width=\textwidth]{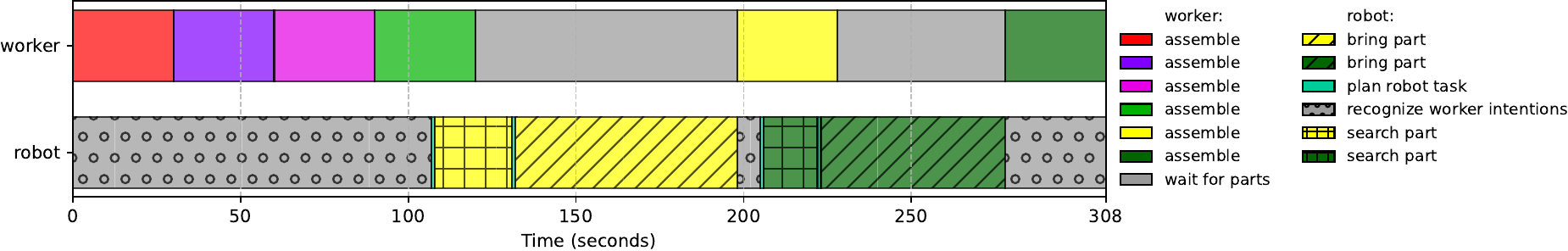}
    \caption{Timeline of an exemplary scenario, where a human worker assembles a type-A insect hotel while the robot perceives the environment and brings necessary parts. The worker assembles several available parts while the robot observes and plans. After enough information is collected to justify the cost of acting, the robot begins searching for and bringing the missing yellow and dark green parts, both needed for this hotel type.
    The colors of ``assemble'', ``search'' and ``bring'' actions match the respective parts, and ``recognize worker intentions'' consists of multiple steps of interleaved observations and planning on the AGR-POMDP.}
    \label{fig:timeline_plot}
\end{figure*}

Figure~\ref{fig:timeline_plot} illustrates an archetypal scenario, where a human worker assembles a type-A hotel and the initial inventory contains red, purple, magenta, orange and bright green parts, while yellow, dark green and black parts are missing. The robot has no initial knowledge about assembly status, inventory or intended hotel type, and must estimate all of these variables.
Establishing a connection between assembled parts and hotel type requires multiple observations and planning steps (``recognize intentions'' in Figure~\ref{fig:timeline_plot}), which may take longer in the beginning due to the size of the underlying AGR-POMDP. After several observations and planning steps, the robot brings the first missing part (yellow, a common part), followed by further observation and planning. At this point enough confidence is reached, via observations, to safely bring dark green (type-A specific) and avoid black (type-B). Our robot often schedules common parts first and waits on type-specific parts, because we adopted a risk-averse approach and reward structure.
It must be noted that the relationship between the worker's plan (type-A), their needs and part status (yellow and dark green missing) emerges from solving the AGR-POMDP, and is never given explicitly through rules or plans in advance. The worker may mix parts or classification errors may exist, and the robot would simply respond with updated information. This proactive, emergent behavior is, however, computationally expensive.

The scenario in Figure~\ref{fig:timeline_plot} was executed 20 times, all completed successfully with a mean time of 344s, where the worker waited for parts for 227s and the robot needed 192s to search and bring parts. The waiting periods are a result of the complex planning steps together with the slow but safe robot operation, which also requires time to navigate, find and retrieve parts.  The worker also performed assembly actions in 30s intervals, since the insect hotel parts are so easily assembled and the domain is ultimately a metaphor for more complex industrial settings. Although all hotels were successfully completed, these runtimes represent a highly experimental platform with many opportunities for algorithmic and practical improvements.

Standard cases (e.g., fig. ~\ref{fig:timeline_plot}) work quite well but some caveats and corner cases remain. For example, if all unique parts were missing and the hotel type couldn't be detected, the robot simply brought one, observed, and if not assembled it brought another type.
A slow or inactive worker yielded little or no assembly progress, and the robot simply brought missing parts regardless of assembly status. 
Another special case occurs if no required parts are missing, so only information gathering and no manipulation actions are needed. Assessing the performance of an AGR assistant robot is clearly no easy task.



Overall, the observed robot behavior in these test cases is particularly promising if we consider that no \emph{ongoing activities} are recognized. Our simplified perception pipeline detects only color-coded parts, their absence and assembly status. This means the robot can only detect worker actions after they occur, not during. While limiting, we argue this strengthens our proposal since it is nevertheless able to approximate and respond to the worker's needs in a highly uncertain environment.

\section{Conclusions and discussion}
Our approach to integrate perception, planning and acting with active goal recognition focuses primarily on the underlying uncertainty in human-robot collaborative scenarios.  The object assembly scenario is surprisingly challenging, but our robot demonstrated flexible and adaptive behavior that balances cost and opportunity, while helping a person achieve goals. Integrating these functions onboard a robot incurs in a significant computational cost, but the highly modular nature of our framework allows us to continue improving all areas independently: from faster function approximation methods to improved robot response times. We are particularly interested in designing more responsive AGR assistants, in application domains such as smart manufacturing and ``Industry 4.0''.
 


\bibliographystyle{IEEEtran}
\bibliography{references}

\end{document}